\documentclass{article}

\usepackage{microtype}
\usepackage{graphicx}
\usepackage{subfigure}
\usepackage{booktabs, amsmath, bbm, amsfonts} 
\usepackage{multirow}
\usepackage{textcomp}
\usepackage{color, colortbl}
\definecolor{Gray}{gray}{0.9}
\usepackage{hyperref}

\usepackage[accepted]{icml2021}

\icmltitlerunning{Few-Example Clustering via Contrastive Learning}
\begin{document}
\twocolumn[
\icmltitle{Few-Example Clustering via Contrastive Learning}



\begin{icmlauthorlist}
	\icmlauthor{Minguk Jang}{kaist}
	\icmlauthor{Sae-Young Chung}{kaist}
\end{icmlauthorlist}

\icmlaffiliation{kaist}{School of Electrical Engineering, Korea Advanced Institute of Science and Technology (KAIST), Daejeon, Korea}

\icmlcorrespondingauthor{Minguk Jang}{mgjang@kaist.ac.kr}
\icmlcorrespondingauthor{Sae-Young Chung}{schung@kaist.ac.kr}
\icmlkeywords{Machine Learning, ICML}

\vskip 0.3in
]


\printAffiliationsAndNotice{\icmlEqualContribution} 

\begin{abstract}
	We propose Few-Example Clustering (FEC), a novel algorithm that performs contrastive learning to cluster few examples. Our method is composed of the following three steps: (1) generation of candidate cluster assignments, (2) contrastive learning for each cluster assignment, and (3) selection of the best candidate. Based on the hypothesis that the contrastive learner with the ground-truth cluster assignment is trained faster than the others, we choose the candidate with the smallest training loss in the early stage of learning in step (3). Extensive experiments on the \textit{mini}-ImageNet and CUB-200-2011 datasets show that FEC outperforms other baselines by about 3.2\% on average under various scenarios. FEC also exhibits an interesting learning curve where clustering performance gradually increases and then sharply drops.
\end{abstract}

\section{Introduction}
\begin{figure*}[ht]
	\vskip -0.1in
	\begin{center}
		\includegraphics[width=1.0\linewidth]{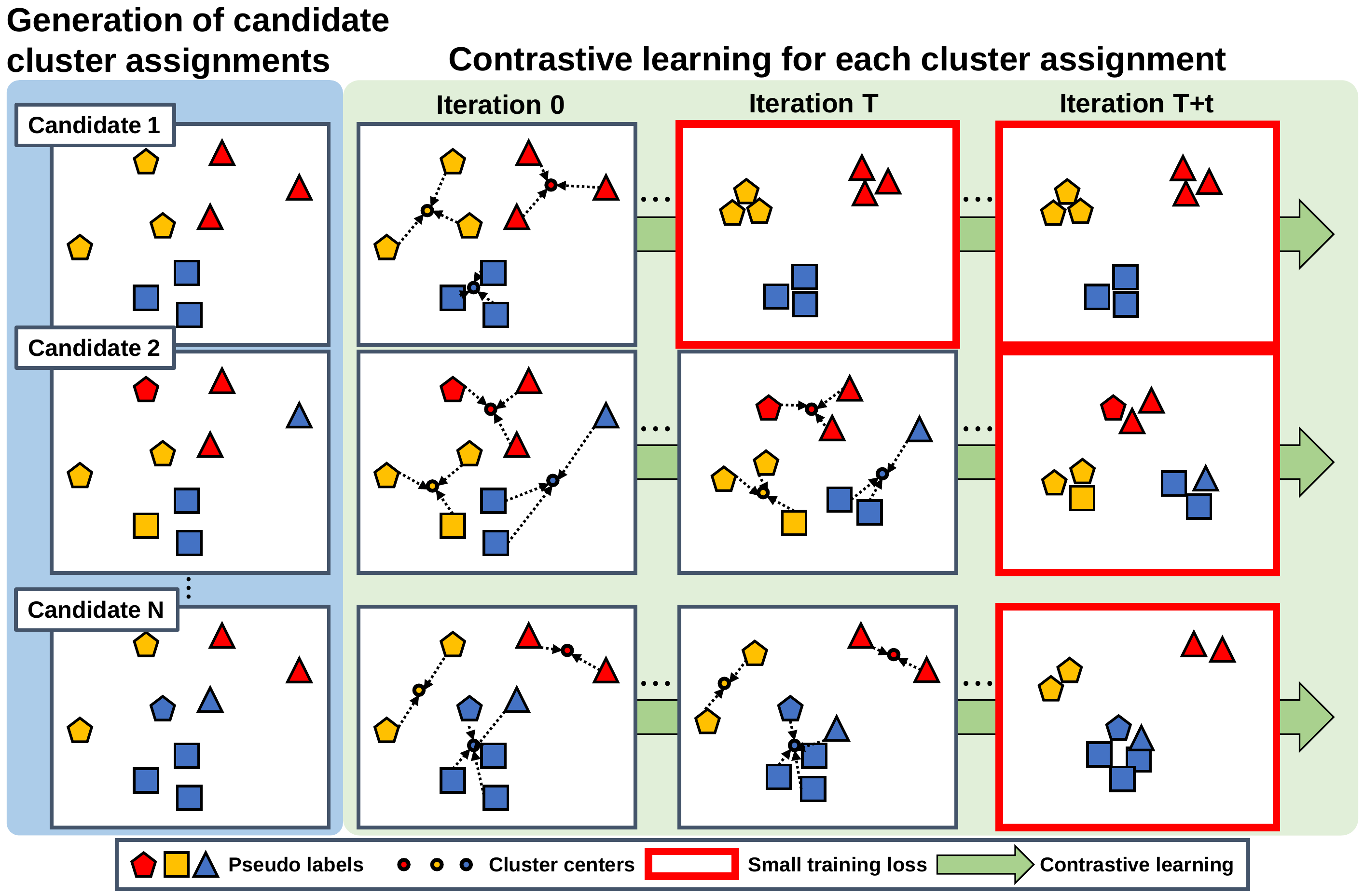}
	\end{center}
	\caption{A conceptual illustration of our proposed algorithm FEC in the feature space. 
		The same shape indicates the same ground-truth cluster.
		For each cluster assignment, we fine-tune the feature network using contrastive learning that minimizes the distance between the examples within the same cluster and maximizes the distance between the examples from different clusters in the feature space.
		We regard the candidate with the smallest training loss in the early stage of learning as the best candidate.
		In this figure, since the candidate 1 converges faster than the other candidates, we choose the candidate 1 as the best one.
	}
	\label{fig:illustration}
	\vskip -0.1in
\end{figure*}
Clustering, which is one of the most popular unsupervised techniques, aims to group similar examples into the same cluster by a similarity measure.
Although conventional clustering methods such as K-means clustering \cite{MacQueen} are powerful on a large-scale dataset, they are ineffective for High Dimension, Low Sample Size (HDLSS) examples. \cite{Ahn, Terada1, Sarkar, Shen}.
Studies on clustering HDLSS data can be helpful for many applications, including neurological diseases analysis \cite{Datta, Yufeng, Alashwal}.

Recently, deep learning techniques have achieved great progress in a variety of areas including visual recognition and language translation \cite{Krizhevsky, Simonyan1, Devlin}. However, to the best of our knowledge, not much progress has been made in utilizing deep neural networks on clustering HDLSS data.

In this paper, we propose \textit{Few-Example Clustering} (FEC), a novel clustering algorithm based on the hypothesis that the contrastive learner with the ground-truth cluster assignment is trained faster than the others. This hypothesis is built on the phenomenon that deep neural networks initially learn patterns from the training examples.
FEC is composed of the following three steps (see Figure~\ref{fig:illustration}): (1) generation of candidate cluster assignments, (2) contrastive learning for each cluster assignment, and (3) selection of the best candidate.
In step (1), we generate candidate cluster assignments using a pre-trained feature network.
In step (2), for each candidate, we fine-tune the feature network using contrastive learning that minimizes the distance between examples in the same cluster and maximizes the distance between examples from different clusters in the feature space.
Based on the hypothesis mentioned earlier, we choose the candidate with the smallest training loss in the early stage of learning in step (3).

We investigate the effectiveness of FEC under various scenarios on the \textit{mini}-ImageNet and CUB-200-2011 datasets.
We evaluate on the task to group five examples into two clusters of sizes one and four. FEC outperforms other baselines by about 3.2\% on average in accuracy. 
We further evaluate on the task to group 80 examples into 5 clusters.
FEC outperforms other baselines by about 0.013 and 0.018 on average in Adjusted Rand Index (ARI) and Normalized Mutual Information (NMI), respectively.

\section{Related works}
\textbf{Training dynamics of DNNs:} 
\citet{Arpit} showed that deep neural networks (DNNs) initially learn patterns from the training examples during training. 
In the presence of label noise, the examples with noisy labels do not share the patterns of the examples with clean labels.
Based on the phenomenon, some studies regard the examples	that have small losses in the initial phase of learning as examples with clean labels \cite{Jiang, Han, Jisoo}.

Inspired by the small-loss criterion to find clean examples, we regard the candidate with the smallest training loss in the early stage of learning as the best candidate in step (3) of our algorithm.

\textbf{Self-supervised learning:}
Our method is related to two self-supervised approaches: deep clustering and contrastive learning.

Deep clustering that combines clustering and representation learning is one of the most promising approaches for self-supervised learning \cite{xie2016unsupervised, Caron, Asano, Ji, Sharma}.
Most studies in this approach iteratively (1) perform clustering methods such as K-means for pseudo-labeling on unlabeled examples and (2) learn representation using supervised learning with the pseudo labels.
Deep clustering has shown comparative representation learning performance to supervised learning.

Contrastive learning aims to learn representations by minimizing the distances between positive (\textit{i.e.}, similar) pairs and maximizing the distance between negative (\textit{i.e.}, dissimilar) pairs.
The positive and negative pairs can be generated by data augmentation \cite{Chen, Grill, Khosla} or they can be identified by clustering methods such as K-means in guided SimCLR \cite{Chakraborty, li2020prototypical, Sharma}.
Recently, contrastive learning methods have shown state-of-the-art performance in self-supervised learning.

\textbf{Few-shot learning:}
Learning in a few-data regime has been largely discussed in the context of classification \cite{Ravi, Chen_closer}.
One direction of few-shot learning research is inductive few-shot learning, which uses labeled examples to adapt a classifier to the novel task.  
Some studies in this direction are based on metric learning. The approaches, which seek to learn embeddings with good generalization ability \cite{Vinyals, Snell}, task-adaptive metric \cite{Oreshkin, Sung}, and task-adaptive embeddings \cite{Yoon, Lichtenstein}.
Other studies in this direction are gradient-based approaches, which seek to find an initialization that facilitates few-shot adaptation \cite{Finn, Nichol, Rusu}.
Another direction of few-shot learning research is transductive few-shot learning, which uses not only labeled examples but also unlabeled examples to adapt a classifier to the novel task.
Most studies in this direction are based on label propagation \cite{Liu, Qiao}.

\textbf{Sinkhorn K-means clustering:}
Sinkhorn K-means clustering \cite{Asano, Genevay, Huang} is a variant of K-means that regards K-means clustering as an optimal transportation problem and finds cluster assignment by using the Sinkhorn-Knopp algorithm under the constraint that cluster sizes are equal. \textit{i.e.}, Sinkhorn K-means clustering aims to cluster $N$ examples $\{x_1,x_2,\cdots, x_N\}$ into $K$ clusters. Sinkhorn K-means to get the assignment $p$ is formulated as follows:
\begin{align}
	\min_{p,\mu} &\sum_{j=1}^K \sum_{i=1}^N  p_{i,j} d(x_i,\mu_j) - \gamma H(p) \\
	\text{s.t.} & \sum_{j=1}^K p_{i,j} = \frac{1}{N}, \forall i\in\{1,\ldots,N\} \\
	& \sum_{i=1}^N p_{i,j} = \frac{1}{K}, \forall j\in\{1,\ldots,K\} \\
	& 0 \le p_{i,j} \le 1,
\end{align}
where $H(p)$ denotes the entropy of the assignment $p$, $\mu_k$ denotes the center of the $k$-th cluster determined by the assignment $p$, and $\gamma$ denotes a hyperparameter for the entropy term.
More details on Sinkhorn K-means clustering will be described in Appendix A.

\section{Methodology}
\begin{algorithm}[!tb]
	\caption{FEC with exhaustive search}
	\label{alg:es}
	\begin{algorithmic}
		\STATE {\bfseries Input:} pre-trained feature model $f_\theta$, unlabeled dataset $\mathcal{D}$, number of clusters $N$, size of each cluster $K$, number of candidates $N_{c}$, softmax temperature $\alpha$
		\STATE Consider all possible cluster assignments as candidates
		\STATE Generate randomly initialized additional layers $g_\phi^{c}$ for each candidate $c$
		\REPEAT
		\FOR{$p=1$ {\bfseries to} $N_{c}$}
		\STATE Train $g_\phi^{c}$ to minimize the loss $\mathcal{L}(\mathcal{D}^c;\phi,\alpha)$ in (\ref{eq:loss})
		\ENDFOR
		\UNTIL{training loss of the best candidate is converged}
		\STATE Choose the candidate $c^*$ that has the smallest training loss.
	\end{algorithmic}
\end{algorithm}

\begin{algorithm}[!tb]
	\caption{FEC with iterative partial search}
	\label{alg:cm}
	\begin{algorithmic}
		\STATE {\bfseries Input:} pre-trained feature model $f_\theta$, unlabeled dataset $\mathcal{D}$, number of clusters $N$, number of candidates $N_c$, duration of fine-tuning $T_{f}$, assignment refinement period $T_{r}$, softmax temperature $\alpha$, clustering method $\mathcal{E}$
		\STATE Generate $N_c$ candidate cluster assignments by running the clustering algorithm $\mathcal{E}$ multiple times in the feature space $f_\theta $
		\STATE Generate randomly initialized additional layers $g_\phi^{c}$ for each candidate $c$
		\FOR{$t=1$ {\bfseries to} $T_{f}$}
		\FOR{$c=1$ {\bfseries to} $N_{c}$}
		\STATE Train $g_\phi^{c}$ to minimize the loss $\mathcal{L}(\mathcal{D}^c;\phi,\alpha)$ in (\ref{eq:loss})
		\IF{$t \equiv -1 \pmod{T_{r}}$}
		\STATE Refine the candidate cluster assignment $c$ by running the clustering algorithm $\mathcal{E}$ in the feature space $ g_\phi^{c} \circ f_\theta $
		\STATE Re-initialize the additional layers $g_\phi^{c}$
		\ENDIF
		\ENDFOR
		\ENDFOR	
		\STATE  Choose the candidate $c^*$ that has the smallest training loss.
	\end{algorithmic}
\end{algorithm}

\subsection{Problem statement}
Suppose that a pre-trained feature model $f_\theta$ is given. Let $\mathcal{D}=\{x_1, x_2, \cdots, x_N\}$ denote a dataset consisting of $N$ unlabeled examples. Our goal is to cluster $N$ examples in the dataset $\mathcal{D}$ into $K$ clusters.

\subsection{FEC with exhaustive search} \label{method:FEC_w_es}
FEC with exhaustive search is composed of three steps as detailed in Algorithm~\ref{alg:es}: (1) generation of candidate cluster assignments, (2) contrastive learning for each cluster assignment, and (3) selection of the best candidate.

\textbf{Generation of candidate cluster assignments:} 
We consider all possible cluster assignments as candidate cluster assignments.

\textbf{Contrastive learning for each cluster assignment:} \label{method:Contrastive learning}
For each candidate, we fine-tune the feature network using contrastive learning that minimizes the distance between examples in the same cluster and maximizes the distance between examples from different clusters in the feature space.

For each candidate cluster assignment $c$, we assign pseudo labels to the unlabeled examples as $\mathcal{D}^c:=\{(x_i, l^c_i)|1\le i \le N\}$, where $l_i^c$ is the pseudo label of $x_i$ determined by the candidate $c$.
To fine-tune feature models using contrastive learning, we freeze the pre-trained network $f_\theta$ and add new trainable layers $g_{\phi}^c$ on top of $f_\theta$.
With a metric $d$, we minimize the distance between an example $x$ and the center of the cluster $x$ belongs to and maximize the distance between the example and the other cluster centers in the feature space of $g_\phi^c \circ f_\theta$. \textit{i.e.}, we minimize the following loss
\begin{align}
	\begin{split}
		\min_{\phi}\;&\mathcal{L}(\mathcal{D}^c;\phi,\alpha)  \label{eq:loss}  \\
		&= \sum_{k=1}^{K} \sum_{(x_i,l_i^c=k) \in \mathcal{D}^c}\frac{\exp(-\alpha d(g_\phi^c \circ f_\theta(x_i), \mu_k^c))}{\sum_j \exp(-\alpha d(g_\phi^c \circ f_\theta(x_i), \mu_j^c))},
	\end{split}
\end{align}
where $\alpha$ denotes the softmax temperature and $\mu_k^c$ denotes the center of the $k$-th cluster determined by the candidate $c$.
In addition, we use an ensemble method to reduce the influences caused by random initializations of additional layers.

\textbf{Selection of the best candidate:}
To select the best candidate, we hypothesize that the contrastive learner with the ground-truth cluster assignment is trained faster than the others. This hypothesis is built on the phenomenon that deep neural networks learn patterns from the training examples in the early stage of training.
Based on the hypothesis, we choose the candidate with the smallest training loss in the early stage of learning. 
Therefore, we need an early stopping criterion.
We terminate learning when the training loss of the best candidate converges (\textit{i.e.} when the decrease in training loss is less than a threshold $\delta$).

\subsection{FEC with iterative partial search} \label{method:FEC_w_pc}
FEC with iterative partial search differs from FEC with exhaustive search in two aspects: (1) FEC with iterative partial search deals with a small subset of all possible cluster assignments and (2) iteratively performs contrastive learning and cluster assignment refinement (see Algorithm~\ref{alg:cm}).

Since the number of all possible cluster assignments exponentially increases as the number of examples increases, FEC with exhaustive search is impossible when the number of examples is large.
Therefore, we only deal with $N_c$ candidate cluster assignments in FEC with iterative partial search.

Deep clustering methods \cite{Caron, Asano, Ji} improve cluster assignment by iteratively performing clustering methods on the learned feature space to generate pseudo labels and adapting representations using supervised learning with the pseudo labels.
Inspired by the iterative updates in deep clustering methods, we refine the candidate cluster assignments on the feature space periodically in FEC with iterative partial search.

Both FEC with iterative partial search and FEC with exhaustive search select the best candidate based on the same hypothesis mentioned earlier.
Thus, FEC with iterative partial search also needs an early stopping criterion.
We tried several early stopping criteria for FEC with iterative partial search, but it was the most effective to terminate after a fixed number of iterations for all tasks.

In addition, we use an ensemble method to reduce the influence caused by random initializations of additional layers. For each cluster assignment, ensemble members are trained in parallel. Even with the same training examples, all members show different clustering outcomes due to random initializations. When we refine the cluster assignment, we choose the ensemble member with the smallest training loss and refine the assignment in the learned feature space of the chosen member. Then, we synchronize all the cluster assignments of members with the refined cluster assignments. The ensemble version of FEC with iterative partial search is described in Appendix D.

\section{Experiments} \label{sec:experiment}
\begin{figure*}[tb]
	\vskip -0.1in
	\begin{center}
		\includegraphics[width=1.0\linewidth]{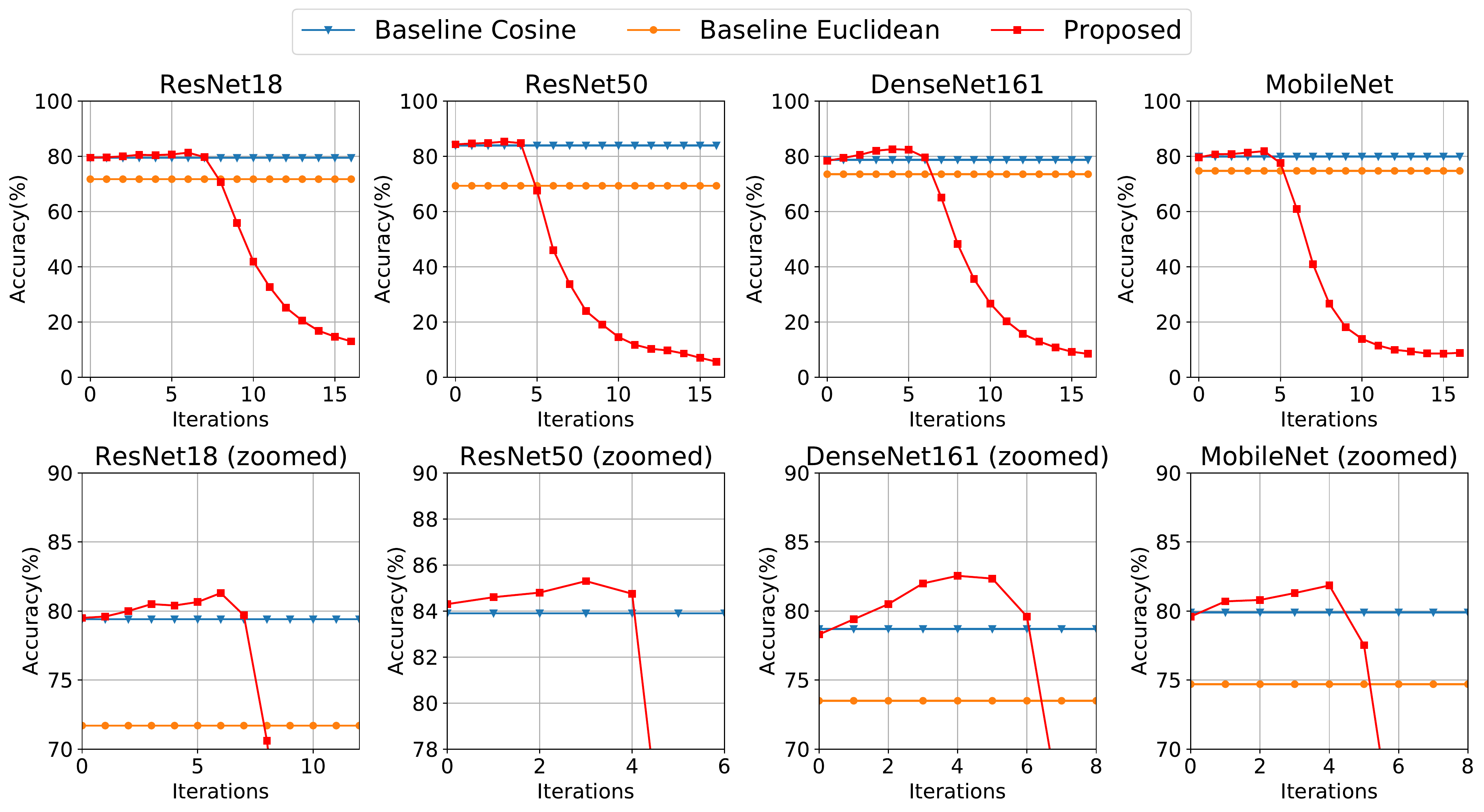}
	\end{center}
	\caption{Training curves of FEC on 4:1 clustering task under the scenario \textit{mini}-ImageNet$\rightarrow$CUB. The clustering accuracy reaches peaks at the early stage of learning. However, as the learning continues, the accuracy sharply drops.}
	\label{fig:41clustering}
	\vskip -0.1in
\end{figure*}

\begin{table*}[tb]
	\vskip 0.15in
	\caption{Average accuracy(\%) on the 4:1 clustering task under different three scenarios: \textit{mini}-ImageNet$\rightarrow$\textit{mini}-ImageNet, \textit{mini}-ImageNet$\rightarrow$CUB, and ImageNet$\rightarrow$CUB. Our proposed FEC outperforms all baselines under vairous scenarios.}
	\label{table:41clustering}
	\begin{center}
		\begin{small}
			\begin{tabular}{l|c|ccc}
				\toprule
				\multirow{2}{*}{Method}& \multirow{2}{*}{Backbone}          & \multicolumn{3}{c}{Accuracy (\%)}            \\
				&  & \textit{mini} $\rightarrow$ \textit{mini} & \textit{mini} $\rightarrow$ CUB & ImageNet $\rightarrow$ CUB \\ \midrule
				Euclidean         & ResNet18  & 62.7        & 47.2        & 71.4        \\
				Cosine            & ResNet18  & 73.5        & 50.0        & 79.4        \\
				PCA+Euclidean     & ResNet18  & 62.7        & 47.8        & 71.7        \\
				PCA+Cosine        & ResNet18  & 73.5        & 51.4        & 79.4        \\
				\rowcolor{Gray}
				\textbf{FEC (Proposed)} & ResNet18  & \textbf{76.9} & \textbf{53.1} & \textbf{82.0}\\ \midrule
				Euclidean         & ResNet50  & 52.0        & 42.4        & 69.3        \\
				Cosine            & ResNet50  & 57.2        & 45.2       & 83.9        \\
				PCA+Euclidean     & ResNet50  & 52.2        & 43.9        & 70.0        \\
				PCA+Cosine        & ResNet50  & 58.7        & 46.2        & 83.9        \\
				\rowcolor{Gray}
				\textbf{FEC (Proposed)} & ResNet50  & \textbf{63.0} & \textbf{48.4}& \textbf{88.9} \\ \midrule
				Euclidean         & DenseNet  & 63.9        & 50.7        & 73.5        \\
				Cosine            & DenseNet  & 74.7        & 53.5        & 78.7        \\
				PCA+Euclidean     & DenseNet  & 64.8        & 50.7        & 73.5        \\
				PCA+Cosine        & DenseNet  & 75.0        & 53.7        & 78.7        \\
				\rowcolor{Gray}
				\textbf{FEC (Proposed)} & DenseNet  & \textbf{75.6} & \textbf{55.4} & \textbf{84.4}\\ \midrule
				Euclidean         & MobileNet & 64.1        & 51.4        & 74.7        \\
				Cosine            & MobileNet & 74.1        & 54.0        & 79.9        \\
				PCA+Euclidean     & MobileNet & 64.1        & 52.2        & 74.8        \\
				PCA+Cosine        & MobileNet & 74.2        & 54.4        & 80.5        \\
				\rowcolor{Gray}
				\textbf{FEC (Proposed)} & MobileNet & \textbf{77.4} & \textbf{56.6}& \textbf{86.8} \\ \bottomrule
			\end{tabular}
		\end{small}
	\end{center}
	\vskip -0.1in
\end{table*}

In this section, we show the effectiveness of FEC under various scenarios on the \textit{mini}-ImageNet and CUB-200-2011 datasets. We test on the task to group five examples into two clusters of sizes one and four. We further test on the few-shot learning task when labeled examples are unavailable.

\subsection{Experimental setup}
\textbf{Datasets:}
We consider two widely used datasets in few-shot learning research: the \textit{mini}-ImageNet \cite{Deng} and CUB-200-2011 \cite{Wah} datasets (hereinafter referred to as CUB).
The \textit{mini}-ImageNet dataset is composed of generic images and used to test generic visual recognition ability.
The CUB dataset is composed of bird images and used to test fine-grained image clustering ability.
We resize images to $84\times84$ for \textit{mini}-ImageNet pre-trained models and to $224\times224$ for ImageNet pre-trained models.
More details on these datasets will be described in Appendix B.

\textbf{Evaluation scenarios:}
We test FEC on three different evaluation scenarios motivated from \cite{Chen_closer}: generic image clustering scenario (\textit{mini}-ImageNet $\rightarrow$ \textit{mini}-ImageNet) and cross-domain clustering scenarios (\textit{mini}-ImageNet $\rightarrow$ CUB and ImageNet $\rightarrow$ CUB).
In the generic image clustering scenario, we cluster generic images from the \textit{mini}-ImageNet dataset using the \textit{mini}-ImageNet pre-trained models.
In the cross-domain clustering scenarios, we cluster fine-grained images from the CUB dataset using the \textit{mini}-ImageNet/ImageNet pre-trained models.
We remark that the examples in the clustering task are not used for pre-training.

\textbf{Architectures:} 
We examine FEC with various architectures of the feature models, ResNet18, ResNet50, DenseNet, and MobileNet, which are widely used for ImageNet classification tasks. 
The additional layers are composed of a single layer and two layers for the experiments in Section~\ref{exper:clustering_w_search} and Section~\ref{exper:clustering_w_methods}, respectively.
The output dimension of the additional layers is set to 512.
More details on these architectures will be described in Appendix C.

\textbf{Hyperparameters:}
FEC with exhaustive search involves three hyperparameters: early stopping threshold $\delta$, softmax temperature $\alpha$, and the number of ensembles $N_{e}$. $\alpha$ is set to 10, $N_{e}$ is set to 32, and $\delta$ is set to $10^{-5}$ for the most experiments and $10^{-3}$ for the experiments using \textit{mini}-ImageNet pre-trained ResNet50. 

FEC with iterative partial search involves four hyperparameters: the period of assignment refinement $T_{r}$ and duration of fine-tuning $T_{s}$, softmax temperature $\alpha$, and the number of ensembles $N_{e}$. $N_{e}$ is set to 5, and $\alpha$ is set to 10.
When we use \textit{mini}-ImageNet pre-trained models, $T_{r}$ and $T_{s}$ are set to 4 and 64, respectively.
When we use ImageNet pre-trained models, $T_{r}$ and $T_{s}$ are set to 8 and 16, respectively.
More experimental results on other combinations of $T_{s}$ and $T_{r}$ are summarized in Appendix E.

\textbf{Evaluation metrics for clustering:}
We use the standard metrics: accuracy, Normalized Mutual Information (NMI), and Adjusted Rand Index (ARI)~\cite{Vinh}. Accuracy measures how accurately we find the ground-truth cluster assignment when we use FEC with exhaustive search. For FEC with iterative partial search, we use NMI and ARI to measure our clustering quality. 

\textbf{Implementation details:}
Unless otherwise specified, we use cosine similarity as a similarity measure in our methods.
For all experiments, the averaged performances of over 1000 tasks are reported.
For pre-trained feature models, we follow the training procedure described in \cite{Wang} and the implementation in PyTorch hub~\cite{Paszke} for \textit{mini}-ImageNet and ImageNet, respectively.
To fine-tune the additional layers, we use Adam \cite{Kingma} as the optimizer with an initial learning rate of $10^{-3}$. 


\begin{figure*}[!tb]
	\vskip -0.1in
	\begin{center}
		\includegraphics[width=1.0\linewidth]{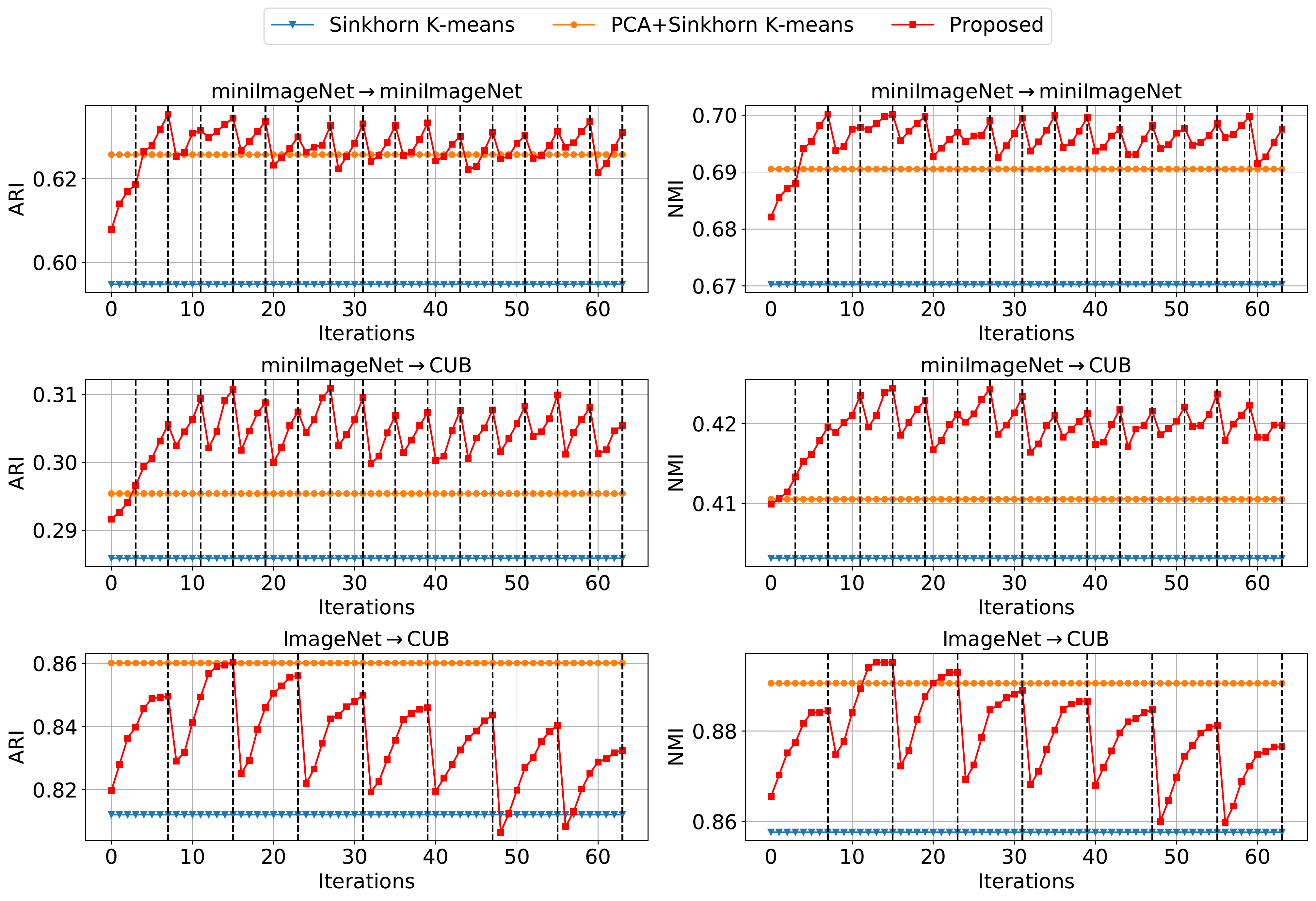}
	\end{center}
	\caption{Training curves of FEC+Sinkhorn K-means on the task of clustering 80 examples into 5 clusters with DenseNet. The black dashed lines represent the timings when we refine the pseudo labels. }
	\label{fig:few_data_clustering}
	\vskip -0.1in
\end{figure*}

\subsection{4:1 clustering} \label{exper:clustering_w_search}

In this problem, each task consists of only five images $\mathcal{D} = \{x_1,x_2,\cdots,x_5\}$ with the ground-truth clusters of sizes one and four. 
4:1 clustering task is identical to finding the farthest example from the other four examples. \textit{i.e.}, we aim to find $x_{\hat{n}}$ that satisfies
\begin{align}
	\hat{n} = \max_{n\in \{1,2,\ldots ,5\}} d\left(h(x_n), \frac{1}{4}\sum_{x\in \mathcal{D}_{\mathcal{T}}-x_n} h(x) \right), 
\end{align}
where $d$ denotes a metric and $h$ denotes a pre-trained feature model. 
Since there are only five clustering candidates, we use Algorithm~\ref{alg:es} for this task.
We consider four baselines by combining two metrics, cosine and Euclidean, and two different feature spaces, a fixed feature space and a dimension reduced feature space by Principal Component Analysis (PCA).
We call these baselines as \textit{cosine}, \textit{Euclidean}, \textit{PCA+cosine}, and \textit{PCA+Euclidean}, respectively.
For baselines with PCA, we tried principal dimensions from 1 to 4, and the best result among them is reported in Table~\ref{table:41clustering}.

The training curves of FEC on the scenario ImageNet $\rightarrow$ CUB can be found in Figure~\ref{fig:41clustering}.  
We can observe that the clustering accuracy reaches peaks at the early stage of learning. 
This observation is consistent with the expectation that the contrastive learner with the ground-truth cluster assignment is learned faster than the others. 
However, as the learning continues, the accuracy sharply drops, which could be due to over-fitting to non-significant patterns.

The overall experimental results on the 4:1 clustering task are summarized in Table~\ref{table:41clustering}. 
We can observe that the baselines using cosine similarity show better performance than the baselines using Euclidean metric, and PCA does not improve performances of baselines Cosine and Euclidean in this task. 
Our proposed algorithm FEC substantially outperforms all baselines by about 3\% on average in accuracy. 

Comparing Figure~\ref{fig:41clustering} and Table~\ref{table:41clustering} shows that using different early stopping for each task increased the clustering performance by about 2\% rather than stopping after a fixed number of iterations for all tasks.

\subsection{Clustering 80 examples into 5 clusters} \label{exper:clustering_w_methods}
In this problem, we aim to group 80 examples into 5 clusters of equal sizes. Since the complexity of this problem is high, we use Algorithm~\ref{alg:cm} instead of Algorithm~\ref{alg:es} for this task.
We consider three baselines based on K-means clustering: \textit{K-means}, \textit{Sinkhorn K-means}, and \textit{PCA+Sinkhorn K-means}.
PCA+Sinkhorn K-means clusters the examples in the task by performing Sinkhorn K-means clustering on the reduced dimension feature space by PCA. 
For PCA+Sinkhorn K-means, we tried principal dimensions of 2, 4, 8, 16, 32, and 64, and the best result among them is reported in Table~\ref{table:few_data_clustering}.

\begin{table*}[!tb]
	\vskip 0.15in
	\caption{Average performance on the task of clustering 80 examples into 5 clusters under three different scenarios: \textit{mini}-ImageNet$\rightarrow$\textit{mini}-ImageNet, \textit{mini}-ImageNet$\rightarrow$CUB, and ImageNet$\rightarrow$CUB. Our proposed FEC+Sinkhorn K-means mostly outperforms all baselines under various scenarios.}
	\label{table:few_data_clustering}
	\begin{center}
		\begin{small}
			\begin{tabular}{l|c|cc|cc|cc}
				\toprule
				\multirow{2}{*}{Method}& \multirow{2}{*}{Backbone}          & \multicolumn{2}{c|}{\textit{mini} $\rightarrow$ \textit{mini}}   & \multicolumn{2}{c|}{\textit{mini} $\rightarrow$ CUB} & \multicolumn{2}{c}{ImageNet $\rightarrow$ CUB}          \\
				&  & ARI & NMI  & ARI & NMI  & ARI & NMI \\ \midrule
				K-means         & ResNet18                            & 0.4654  & 0.5888  & 0.2502 & 0.3810  & 0.6128 & 0.7372      \\
				Sinkhorn K-means           & ResNet18                 & 0.5690 & 0.6503 & 0.2779   & 0.3952 & 0.7891 & 0.8358     \\
				PCA+Sinkhorn K-means     & ResNet18                   & 0.5978  & 0.6683      & 0.2840 & 0.3993 & \textbf{0.8159} & 0.8538     \\
				\rowcolor{Gray}
				\textbf{FEC+Sinkhorn K-means (Proposed) }  & ResNet18  & \textbf{0.6075} &  \textbf{0.6781} & \textbf{0.2963} & \textbf{ 0.4137 }& 0.8155 &  \textbf{0.8607} \\ \midrule
				K-means         & ResNet50  & 0.3322       & 0.4705      & 0.2116 & 0.3354 & 0.6623 & 0.7811      \\
				Sinkhorn K-means             & ResNet50  & 0.3366      & 0.4602        &0.2231   &0.3400 &0.8229&  0.8655     \\
				PCA+Sinkhorn K-means     & ResNet50  & 0.3659     & 0.4819       & 0.2275 & 0.3428 & 0.8523 & 0.8845      \\
				\rowcolor{Gray}
				\textbf{FEC+Sinkhorn K-means (Proposed) } & ResNet50  & \textbf{0.4405} & \textbf{0.5363}& \textbf{0.2516}  & \textbf{0.3623 }& \textbf{0.8715} & \textbf{0.8992} \\ \midrule
				K-means         & DenseNet  & 0.4791    & 0.6009       & 0.2544  & 0.3844 & 0.5937 & 0.7312   \\
				Sinkhorn K-means            & DenseNet  & 0.5948 & 0.6703    & 0.2859  & 0.4032  & 0.8122 & 0.8576  \\
				PCA+Sinkhorn K-means     & DenseNet  & 0.6258    & 0.6905 & 0.2954   & 0.4105 & 0.8602 & 0.8906   \\
				\rowcolor{Gray}
				\textbf{FEC+Sinkhorn K-means (Proposed) } & DenseNet  & \textbf{0.6311} & \textbf{0.6976}  & \textbf{0.3055} & \textbf{0.4198} & \textbf{0.8605 }& \textbf{0.8952} \\ \midrule
				K-means         & MobileNet & 0.4487  & 0.5677       & 0.2576   & 0.3870 & 0.6082 & 0.7346    \\
				Sinkhorn K-means            & MobileNet & 0.5295       & 0.6172       & 0.2851   & 0.4022& 0.8132&0.8538    \\
				PCA+Sinkhorn K-means     & MobileNet & 0.5588    & 0.6341     &0.2927     & 0.4075 & 0.8475  & 0.8779    \\
				\rowcolor{Gray}
				\textbf{FEC+Sinkhorn K-means (Proposed) } & MobileNet & \textbf{0.5652} & \textbf{0.6436} & \textbf{0.3099}&\textbf{0.4236}&\textbf{0.8483}&\textbf{0.8827}  \\ \bottomrule
			\end{tabular}
			
		\end{small}
	\end{center}
	\vskip -0.1in
\end{table*}

\begin{table*}[tb]
	\vskip 0.15in
	\caption{Average performance on the task of clustering 80 examples into 5 clusters under the scenario \textit{mini}-ImageNet$\rightarrow$\textit{mini}-ImageNet with ResNet50. Select the best candidate: the selection of the candidate with the smallest training loss. Refine assignments: the candidate cluster assignment refinements. Re-initialization: re-initialization of the additional layers.}
	\label{table:ablation}
	\begin{center}
		\begin{small}
			\begin{tabular}{ccc|cc|cc|cc}
				\toprule
				\multirow{2}{*}{Select the best candidate} & \multirow{2}{*}{Refine assignments} & \multirow{2}{*}{Re-initialization} & \multicolumn{2}{c|}{\textit{mini}$\rightarrow$\textit{mini}}&\multicolumn{2}{c|}{\textit{mini}$\rightarrow$CUB}&\multicolumn{2}{c}{ImageNet$\rightarrow$CUB} \\ 
				&         &         & ARI & NMI  & ARI & NMI  & ARI & NMI \\ \midrule
				&         &         & 0.3366 & 0.4602 & 0.2231  &0.3400&0.8229&0.8655 \\
				& $\surd$ &         & 0.3724 & 0.4784 & 0.2206  & 0.3348 & 0.7887 &  0.8465\\
				& $\surd$ & $\surd$ & 0.4204  & 0.5198& 0.2415     &0.3540&0.7870&0.8459 \\ \midrule
				$\surd$ &         & &0.3267  & 0.4526    & 0.2211 &0.3396&0.8523&0.8855  \\
				$\surd$ & $\surd$ &         & 0.4029 & 0.5017 & 0.2342 & 0.3452 & \textbf{0.8715} & 0.8966 \\
				$\surd$ & $\surd$ & $\surd$ & \textbf{0.4405}  & \textbf{0.5363} & \textbf{0.2516} &\textbf{0.3623}&\textbf{0.8715}&\textbf{0.8992}  \\ \bottomrule
			\end{tabular}
		\end{small}
	\end{center}
	\vskip -0.1in
\end{table*}

\subsubsection{Clustering results} 
The training curves of FEC+Sinkhorn K-means with DenseNet can be found in Figure~\ref{fig:few_data_clustering}.
We can observe the training curves of FEC+Sinkhorn K-means oscillate after some iterations, which is due to iterative updates in Algorithm~\ref{alg:cm}.
In the experiments using the ImageNet pre-trained models, we can observe that the clustering performance of FEC+Sinkhorn K-means reaches peaks at the end of the second period, so we early terminate after two periods in such experiments.
However, in the experiments using the \textit{mini}-ImageNet pre-trained models, finding the early termination point is hard, thus we report the performance of FEC+Sinkhorn K-means at the end of learning in Table~\ref{table:few_data_clustering}.

Overall experimental results are summarized in Table~\ref{table:few_data_clustering}. 
Although PCA improves the performance of Sinkhron K-means by about 0.024 and 0.015 on average in ARI and NMI, FEC+Sinkhorn K-means performs even better.
Our proposed algorithm FEC+Sinkhorn K-means outperforms all baselines by about 0.013 and 0.018 on average in ARI and NMI under various scenarios, respectively.

\subsubsection{Ablation study} \label{exper:ablation}

In this section, we study the impact on the performance improvement of the best candidate selection, refinement of candidate cluster assignment, and re-initialization of the additional layers. Results on ResNet50 are summarized in Table~\ref{table:ablation}.

\textbf{Without selecting the best candidate:}
In FEC with iterative partial search, we choose the best candidate that incurs the smallest training loss after a fixed number of iterations.
To understand the effect of the best candidate selection, we ablate the best candidate selection from FEC+Sinkhorn K-means and instead use the average performances of all candidates, which leads to degraded performances as shown in Table~\ref{table:ablation}.

\textbf{Without refining candidate cluster assignments:}
In FEC with iterative partial search, we periodically refine the candidate cluster assignments. 
To study the effect of this, we run FEC+Sinkhorn K-means without the refinement, which shows poor performance as shown in Table~\ref{table:ablation}.

\textbf{Without re-initialization of the additional layers:}
Conventional deep clustering algorithms do not re-initialize the models after the refinement of cluster assignments.
However, we found that FEC+Sinkhorn K-means performs better if we re-initialize the additional layers for fine-tuning.
The results of FEC+Sinkhorn K-means without re-initialization of the additional layers are summarized in Table~\ref{table:ablation}.

\section{Conclusion}

Learning in a few-data regime have been mainly studied in the context of classification, not clustering. Few-example clustering is expected to be helpful for many real-world problems where data collection and labeling are difficult. Nevertheless, conventional clustering methods are ineffective for few-example clustering.
In this paper, we introduce a novel clustering algorithm Few-Example Clustering, which generates candidate cluster assignments using a feature model, fine-tunes the feature model using contrastive learning, and selects the best candidate that incurs the smallest training loss in the early stage of learning.
We experimentally show that our algorithm consistently outperforms other baselines in all few-example clustering tasks we considered.

\bibliography{reference}
\bibliographystyle{icml2021}

\clearpage
\appendix
\section{Sinkhorn K-means} \label{appendix:Sinkhorn K-means}
\textbf{Sinkhorn K-means clustering}
K-means clustering is a method to cluster $N$ examples $\{x_i\}_{i=1,\ldots, N}$ into $K$ clusters in which each example belongs to the cluster with the nearest center. 
With a metric $d$, the objective to get the assignment $p$ is 
\begin{align}
	\min_{p,c} &\sum_{j=1}^K \sum_{i=1}^N  p_{i,j} d(x_i,c_j) \\
	\text{s.t.} & \sum_{j=1}^K p_{i,j} = 1, \forall i\in\{1,\ldots, N\} \\
	& p_{i,j} \in \{0, 1\},
\end{align}
where $c_j$ denote the center of the $j$-th cluster and $p_{i,j}$ denote the assignment example $x_i$ to $j$-th cluster. 
On the other hand, Sinkhorn K-means clustering \cite{Asano, Genevay, Huang} is a variant of K-means that finds the assignment by replacing the minimization problem of K-means into the Sinkhorn-Knopp algorithm-based optimal transportation problem under the constraint that each cluster has an equal size.
The objective of Sinkhorn K-means is 
\begin{align}
	\min_{p,c} &\sum_{j=1}^K \sum_{i=1}^N  p_{i,j} d(x_i,c_j) - \gamma H(p) \\
	\text{s.t.} & \sum_{j=1}^K p_{i,j} = \frac{1}{N}, \forall i\in\{1,\ldots,N\} \\
	& \sum_{i=1}^N p_{i,j} = \frac{1}{K}, \forall j\in\{1,\ldots,K\} \\
	& 0 \le p_{i,j} \le 1,
\end{align}
where $H(p)$ denote the entropy of the assignment $p$, and $\gamma$ is a hyperparameter for the entropy term.

\section{Datasets} \label{appendix:datasets}
We consider two widely-used datasets in few-shot learning : the \textit{mini}-ImageNet \cite{Deng} and the CUB-200-2011 \cite{Wah} (hereinafter referred to as CUB). We use the \textit{mini}-ImageNet dataset to test generic visual object recognition capabilities. The \textit{mini}-ImageNet dataset consists of 100 classes of 600 images each. The \textit{mini}-ImageNet dataset's classes are divided into base, validation, and novel classes, which contains 64, 16, and 20 classes respectively, as described in \cite{Ravi}. We use the CUB dataset to test fine-grained image classification. The CUB dataset contains 11788 images of birds for 200 classes in total. The CUB dataset's classes are divided into 100 base, 50 validation, and 50 test classes following. We resize the images to 84$\times$84 for \textit{mini}-ImageNet trained models, and 224$\times$224 for ImageNet trained models. 

\section{Architectures of feature models} \label{appendix:models}
We evaluate our method using four different convolutional-netowrk architectures.
\begin{itemize}
	\item Residual networks (ResNet-18/50) : We use the standard 18/50-layer architecture. For \textit{mini}-ImageNet pre-trained model, we remove the first two down-sampling and we change the first convolutional layer to use a kernel of size $3\times3$ (rather than $7 \times 7$) pixels since we resize the image to $84\times84$.
	\item Dense convolutional networks (DenseNet-121/161) : We use the standard 121/161-layer architeture. However, for \textit{mini}-ImageNet pre-trained model, we remove the first two down-sampling layers (\textit{i.e.}, we set their stride to 1) and change the first convolutional layer to use a kernel of size $3\times3$ (rather than $7\times7$) pixels.
	\item MobileNet : We use the standard architecture for the ImageNet-pretrained model. However, for \textit{mini}-ImageNet pre-trained model, we remove the first two down-sampling layers from the network.
\end{itemize}
Before clustering, we require pre-trained models. For \textit{mini}-ImageNet pre-trained models, all feature models are trained for 90 epochs with stocahstic gradient descent with batch size of 256. We follow the implementation details described in \cite{Wang}.
For ImageNet pre-trained models, we download the the ImageNet pre-trained models from the PyTorch hub~\cite{Paszke}.

We download ImageNet pre-trained models from the PyTorch hub~\cite{Paszke}.
To obtain feature models trained on \textit{mini}-ImageNet dataset, we train all feature models following the training implementations described in \cite{Wang}.

ImageNet pre-trained models, 

\section{Ensemble version of FEC with clustering methods} \label{appendix:pseudo_algorithms}
\begin{algorithm}[!tb]
	\caption{Ensemble version of FEC with clustering method}
	\label{alg:ens_cm}
	\begin{algorithmic}
		\STATE {\bfseries Input:} pre-trained feature model $f_\theta$, unlabeled dataset $\mathcal{D}$, number of clusters $N$, number of clustering candidates $N_c$, number of learning steps $T_{step}$, pseudo label refinement period $T_{r}$, softmax temperature $\alpha$, number of ensemble $N_{ens}$
		\STATE Generate $N_p$ candidates by running a clustering algorithm multiple times in the space of embedding $f_\theta $
		\STATE Generate $N_{ens}$ randomly initialized feature transformer $g_\phi^{p}$ for each candidate $\mathcal{C}_p$
		\FOR{$s=1$ {\bfseries to} $T_{step}$}
		\FOR{$p=1$ {\bfseries to} $N_{c}$}
		\STATE Train $g_\phi^{p}$ to minimize the loss $\mathcal{L}(\mathcal{D}^p_{\mathcal{T}};\phi,\alpha)$
		\IF{$s \equiv -1 \pmod{T_{r}}$}
		\STATE Refine the clustering candidate $\mathcal{C}_p$ by running a clustering algorithm in the feature space of $ g_\phi^{p} \circ f_\theta $
		\STATE Re-initialize the feature transformer $g_\phi^{p}$
		\ENDIF
		\ENDFOR
		\ENDFOR	
		\STATE Choose the best candidate $\mathcal{C}_{p^*}$ that has the smallest training loss.
	\end{algorithmic}
\end{algorithm} 

\section{Results on the combinations of $T_s$ and $T_r$} \label{appendix:termination_few_shot}

\end{document}